\documentclass[11pt,a4paper]{article}

\usepackage[T1]{fontenc}
\usepackage[utf8]{inputenc}
\usepackage{lmodern}
\usepackage[margin=25mm]{geometry}
\usepackage{amsmath,amssymb}
\usepackage{graphicx}
\usepackage{booktabs,tabularx,array}
\usepackage{algorithm}
\usepackage{algpseudocode}
\usepackage{tikz}
\usetikzlibrary{arrows.meta,positioning,calc}
\usepackage{microtype}
\usepackage{caption}
\usepackage{placeins}
\usepackage[hidelinks]{hyperref}
\usepackage{url}
\renewcommand{\arraystretch}{1.15}

\newcolumntype{Y}{>{\raggedright\arraybackslash}X}
\DeclareMathOperator{\softmax}{softmax}
\DeclareMathOperator{\GroupNorm}{GroupNorm}
\DeclareMathOperator{\DWConv}{DWConv}

\title{OCBI: Order-Sensitive Cross-Branch Interaction\\
for Image Manipulation Localization}
\author{Zhouli Yu}
\date{}

\begin{document}
\maketitle

\begin{abstract}
Image manipulation localization combines semantic structure with forensic
traces to identify modified pixels. Within-branch spatial mixing and
RGB-guided cross-branch mixing can produce different messages when
composed in opposite orders. We propose OCBI, an extension of
Co-Transformers that models the discrepancy between these two composition
orders for localization. A local cross-branch fusion module uses the
commutator of RGB-guided cross-attention and forensic self-attention to
add a learnable signed correction to feature transfer. Pixelwise adaptive
multiscale aggregation then combines spatial contexts before prediction.
OCBI uses the Co-Transformers backbone and supervision. We compare
OCBI with retrained Co-Transformers using CASIA-only
task training, a 50-epoch budget, and final-epoch evaluation. Across five
clean test datasets, OCBI achieves a macro PixelF1 of 0.5773, exceeding
the baseline by 0.43 percentage points. Under JPEG compression,
Gaussian blur, and resizing on three datasets, its average PixelF1
over 21 dataset--transformation conditions exceeds the baseline by
0.95 percentage points.
\end{abstract}

\section{Introduction}

Image manipulation localization identifies the modified pixels in an image.
A visually plausible edit may preserve object appearance while
altering noise patterns, compression artifacts, or local processing
statistics. Conversely, authentic boundaries and textured regions may
resemble manipulation artifacts. Effective localization therefore requires
combining semantic structure with forensic evidence while distinguishing
manipulation traces from ordinary image variation.

Learned forensic representations capture complementary aspects of this evidence.
Constrained convolution suppresses image content while learning
manipulation-sensitive features~\cite{ref2}. Noiseprint extracts camera-model
fingerprints~\cite{ref6}, while forensic similarity learning compares traces
across image regions~\cite{ref16}. The RGB-and-noise detector of Zhou
et al.~\cite{ref24} and CAT-Net~\cite{ref14} combine appearance with
complementary forensic cues. TruFor~\cite{ref9} combines RGB features with
learned forensic fingerprints and estimates prediction reliability.
Co-Transformers~\cite{ref23} models collaboration between semantic
structure and microscopic traces. These studies motivate examining how
semantic guidance affects the spatial propagation of forensic information.

The order of spatial mixing offers another way to examine this interaction.
Within-branch mixing
aggregates positions according to relationships among forensic features,
whereas cross-branch mixing transfers forensic information using
RGB-guided relationships. Aggregating forensic evidence before this
transfer need not produce the same message as transferring it first and
then applying noise-derived spatial mixing. For manipulation localization,
this distinction may matter when semantic boundaries and forensic
discontinuities do not coincide: the two compositions may distribute weak
traces differently across neighboring positions. We test the hypothesis
that their difference can provide a useful correction to cross-branch
transfer for pixelwise manipulation localization.

Building on Co-Transformers, we introduce OCBI to make this composition
difference explicit. At each hierarchical stage, the cross-branch fusion
module, CBFI, constructs two operators on an aligned local spatial grid:
$A$ aggregates forensic values using RGB queries and forensic keys,
while $N$ aggregates them using forensic queries and keys.
For forensic values $V_n$, $A(NV_n)$ applies forensic mixing before
RGB-guided mixing, whereas $N(AV_n)$ applies the same operators in
the reverse order. The commutator $C=AN-NA$ produces their signed
message difference, which is scaled by a learnable coefficient and
added to the cross-branch message before updating the RGB features.
The adaptive multiscale aggregation module, AMSA, then weights spatial contexts
at each pixel to accommodate variation in the extent of manipulated
regions. The method uses the Co-Transformers backbone, forensic extractor,
prediction head, and supervision.

We evaluate OCBI against retrained
Co-Transformers using the same CASIA-only task-training split and
50-epoch budget. On five clean test datasets, OCBI improves PixelF1
on every dataset and raises the macro score by 0.43 percentage points.
Under JPEG compression, Gaussian blur, and resizing, OCBI raises the
average PixelF1 across the 21 tested dataset--transformation conditions
by 0.95 percentage points.

\section{Related Work}

\subsection{Forensic Representations and Complementary Evidence}

Early learning-based localization methods focused on processing traces that
conventional semantic features can overlook. Bayar and Stamm~\cite{ref2}
introduced a constrained convolutional layer for universal manipulation
detection. Bappy et al.~\cite{ref1} combined resampling features, recurrent
modeling, and an encoder--decoder architecture. Camera-based CNN
features~\cite{ref3} and Noiseprint~\cite{ref6} capture camera-related traces.
Learned self-consistency~\cite{ref12} and forensic similarity~\cite{ref16}
use relationships between image regions to support authenticity analysis.

Multibranch models combine these traces with image appearance. The RGB-and-noise
detector of Zhou et al.~\cite{ref24} fuses appearance and noise features to
detect manipulated regions. CAT-Net~\cite{ref14} combines RGB information with
compression evidence in the DCT domain. MVSS-Net~\cite{ref5} uses multiview
feature learning and multiscale supervision, while TruFor~\cite{ref9} combines
RGB information with a learned noise-sensitive fingerprint and estimates
prediction reliability. Co-Transformers~\cite{ref23} models collaboration
between semantic structure and microscopic traces. OCBI extends this
architecture by modeling the difference between two compositions of
local mixing operators.

Semantic Manipulation Localization~\cite{ref31} studies subtle,
meaning-altering edits and introduces a fine-grained benchmark with
pixel-level annotations. This preprint focuses on semantic changes,
complementing methods based on low-level forensic traces.

\subsection{Spatial Interaction and Dense Localization}

Dense localization requires spatial reasoning in addition to forensic
feature extraction. Existing manipulation localization methods obtain
spatial evidence primarily from local anomalies, boundary cues, or
source--target relationships. ManTra-Net~\cite{ref21} traces manipulation
artifacts and detects local anomalies; SPAN~\cite{ref11} organizes patch
relationships through a local attention pyramid.
MFCN~\cite{ref18} jointly models spliced regions and boundaries, while
GSR-Net~\cite{ref25} strengthens boundary cues using generated samples.
BusterNet~\cite{ref20} localizes
source and target regions in copy-move forgeries.

Related dense prediction tasks also use spatial relationships.
Co-saliency models strengthen consistency among related
regions through co-attention~\cite{ref33} and group semantics
propagation~\cite{ref34}. In video saliency, triplet relationship
aggregation~\cite{ref35} extends interaction modeling to multidimensional
feature relationships. These methods model relationships among regions,
groups, or frames. OCBI models composition-order differences between
local semantic and forensic interactions and retains the baseline
edge supervision.

\subsection{Hierarchical Features and Multiscale Aggregation}

Hierarchical representations enable information exchange across spatial
resolutions. SegFormer~\cite{ref22} provides a hierarchical Transformer
encoder. Swin Transformer~\cite{ref15} demonstrates window-based attention
and cross-window communication through shifted windows.
DANet~\cite{ref8} models spatial and channel dependencies for dense
prediction, while DeepLabv3+~\cite{ref4} combines atrous spatial processing
with encoder--decoder design.

Transformer-based representation exchange also appears in multimodal
learning. UTC~\cite{ref27} uses a unified Transformer for visual dialog and
couples answer ranking and generation through inter-task contrastive
learning. In video generation, DiVA~\cite{ref30} injects visual conditioning
features through decoupled cross-attention. OCBI uses attention to exchange semantic and forensic features
for pixelwise localization. CBFI operates within local windows, and AMSA
weights dilated responses before the DANet prediction head.

\subsection{Datasets and Evaluation Protocols}

CASIA~\cite{ref7} and COVERAGE~\cite{ref19} provide benchmarks for image
tampering and copy-move forgery, respectively.
Columbia~\cite{ref10} and IMD2020~\cite{ref17} add splicing and more broadly
annotated manipulation data. Fantastic Reality~\cite{ref13} further
extends the available image-splice data.
These benchmarks differ in manipulation types, imaging conditions, and
annotation settings.
IMDL-BenCo~\cite{ref26} addresses inconsistencies in implementation and
evaluation through a modular benchmark and codebase. We use its data-loading
and metric infrastructure and report both per-dataset scores and an
equally weighted macro-average. Comparisons use a retrained baseline to
match the training split and evaluation procedure.

\section{Method}

\subsection{Problem Formulation and Overall Architecture}

Given an RGB image $I$ and a binary manipulation mask $y$, the model predicts
a pixelwise manipulation probability map $p$ and an auxiliary edge output.
The input tensor has batch size $B$, three channels, and spatial dimensions
$H\times W$. Edge-supervision targets are generated from the manipulation
mask using the baseline mask-construction pipeline.

\begin{figure}[tbp]
  \centering
  \includegraphics[width=\linewidth]{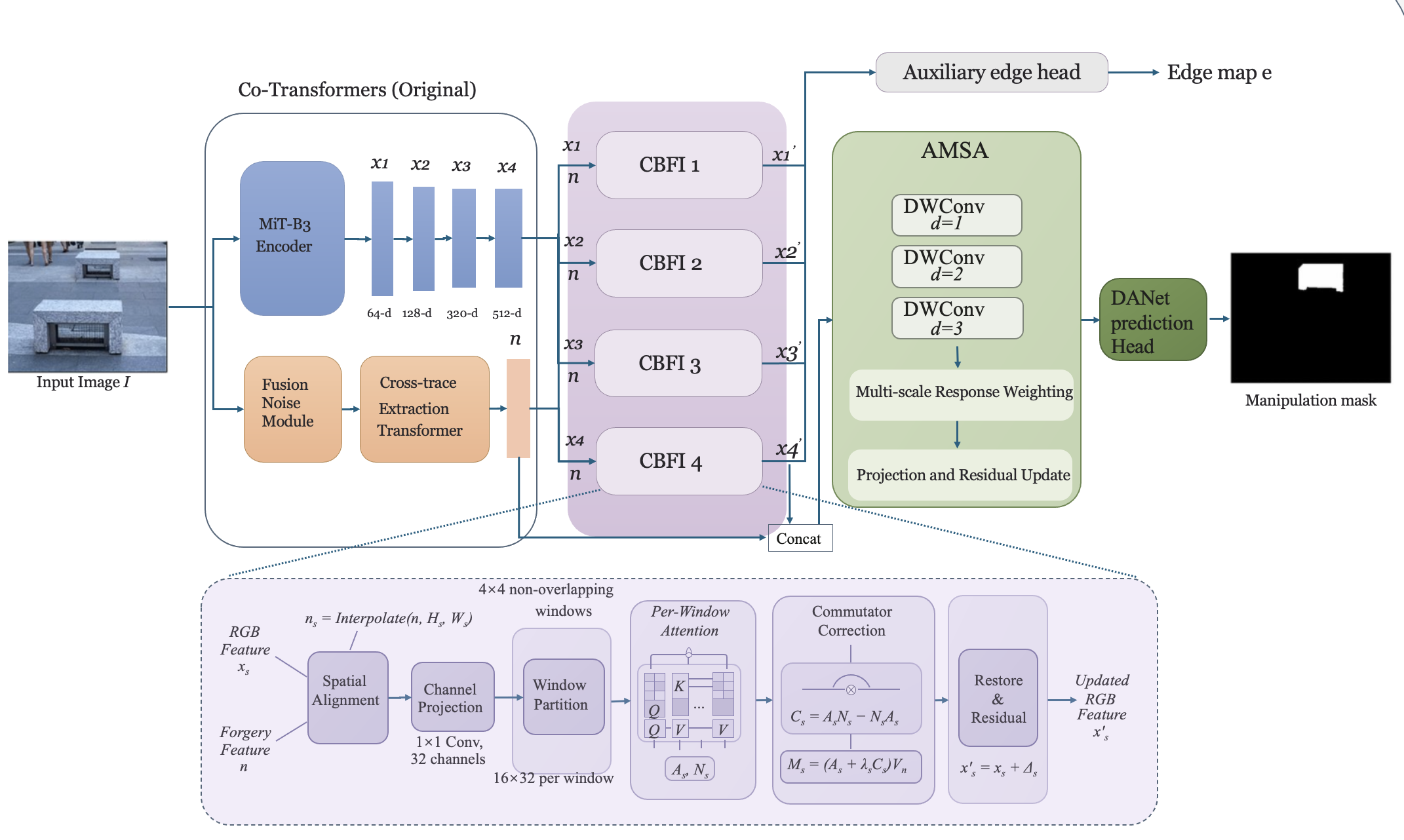}
  \caption{Overall OCBI framework. The Co-Transformers base
  provides the MiT-B3 image encoder, forensic branch, and DANet prediction
  head. Four CBFI modules independently update the encoder outputs using
  aligned forensic features. All four updated features feed the auxiliary
  edge head; only $x'_4$ is concatenated with the forensic feature $n$
  before AMSA and mask prediction. AMSA applies pixelwise weighting to depthwise
  convolution responses at dilation rates 1, 2, and 3, followed by
  projection and residual update. The lower panel details the window-local
  signed interaction in CBFI. The auxiliary edge map supplies supervision
  during training and is not an input to the mask head.}
  \label{fig:architecture}
\end{figure}

The base system is Co-Transformers~\cite{ref23}. Its architecture
contains a hierarchical edge-supervised Transformer, a noise fusion module,
and a Cross-trace Extraction Transformer. We denote the output of the
forensic branch by $n$ and the hierarchical image features by $x_s$. For
$s=1,2,3,4$, the stage channel widths are 64, 128, 320, and 512, respectively.
The encoder uses the MiT-B3 hierarchy of SegFormer~\cite{ref22}.
Figure~\ref{fig:architecture} summarizes the Co-Transformers base
and the OCBI additions.

\FloatBarrier

The MiT-B3 encoder first computes the stage features sequentially:
$x_s=E_s(x_{s-1})$ for $s=1,\ldots,4$, with $x_0=I$.
After all four features have been extracted, independent CBFI modules
produce $x'_s=\mathrm{CBFI}_s(x_s,n)$, preserving each feature's spatial
and channel dimensions. No updated feature $x'_s$ is passed to the next
encoder stage. The auxiliary edge head receives all four updated
features, whereas the mask path uses only $x'_4$:
\begin{equation}
  \begin{aligned}
    \ell_e &= H_{\mathrm{edge}}(x'_1,x'_2,x'_3,x'_4),\\
    z &= \operatorname{Concat}(x'_4,n),\\
    \ell_m &= H_{\mathrm{mask}}\bigl(\operatorname{AMSA}(z)\bigr).
  \end{aligned}
  \label{eq:feature-routing}
\end{equation}
Here, $H_{\mathrm{mask}}$ is the DANet prediction head~\cite{ref8},
and $z$ has 1280 channels. The edge and mask logits, $\ell_e$ and $\ell_m$,
are separately resized to the input resolution and passed through sigmoid.
The edge output is not supplied to the mask head.

CBFI-1--3 contribute through auxiliary edge supervision during training:
the edge loss backpropagates through them into the shared image encoder
and forensic branch, whose learned representations also support mask
prediction. Their corrected features have no direct forward path to the
current mask output. The unmodified features $x_1,x_2,x_3$ still contribute
to $x_4$ through the encoder hierarchy; their CBFI corrections do not.
CBFI-4 participates directly in both the edge and mask paths.

\subsection{Window-Local Cross-Branch Fusion}

At each stage, the forensic feature is first aligned to the spatial
resolution of the image feature by bilinear interpolation. Independent
$1\times1$ convolutions then project both representations to 32 channels.
Related alignment strategies have been used in cross-modal
retrieval~\cite{ref28} and video feature modeling~\cite{ref32}.
CBFI partitions the aligned features into nonoverlapping $4\times4$
windows, each containing $L=16$ positions with 32 channels.

Within a window, $Q_{\mathrm{rgb}}$ and $Q_n$ denote the image and noise
queries, while $K_n$ and $V_n$ denote the noise keys and values. Each matrix
has 16 rows and 32 columns. A hat denotes unit $\ell_2$ normalization along
the feature dimension. Row-wise softmax produces two operators over the
same aligned positions:
\begin{equation}
  A_s=\softmax_{\mathrm{row}}
  \left(\tau_s\widehat{Q}_{\mathrm{rgb}}\widehat{K}_n^{\mathsf T}\right),
  \label{eq:cross-attention}
\end{equation}
\begin{equation}
  N_s=\softmax_{\mathrm{row}}
  \left(\tau_s\widehat{Q}_n\widehat{K}_n^{\mathsf T}\right).
  \label{eq:noise-attention}
\end{equation}
Here, $A_s$ uses image queries and noise keys, whereas $N_s$ uses noise
queries and the same noise keys. Both operators have shape $16\times16$.
Each stage has an independent learnable log-temperature $t_s$, with
$\tau_s=\exp(t_s)$ initialized to an effective temperature of 10.

The order-sensitive message is defined as
\begin{equation}
  C_s=A_sN_s-N_sA_s,\qquad
  M_s=(A_s+\lambda_s C_s)V_n.
  \label{eq:commutator}
\end{equation}
Matrix products act on $V_n$ from right to left:
$A_s(N_sV_n)$ first aggregates forensic values with $N_s$, then applies
the RGB-guided operator $A_s$. Conversely, $N_s(A_sV_n)$ first applies
$A_s$, then mixes the resulting message with the same forensic-derived
weights $N_s$ on the aligned spatial grid. Both compositions use the
same $A_s$ and $N_s$; attention weights are not recomputed between
the two operations. The learnable
coefficient $\lambda_s$ is initialized to 0.5. Although the surrounding
computation uses fp16, attention matrix products and commutator subtraction
are computed in float32.

The correction is signed. Let $\mathbf{1}$ denote the all-ones vector.
Row normalization gives
\begin{equation}
  A_s\mathbf{1}=N_s\mathbf{1}=\mathbf{1},\qquad
  C_s\mathbf{1}=\mathbf{0},\qquad
  (A_s+\lambda_s C_s)\mathbf{1}=\mathbf{1}.
  \label{eq:row-sums}
\end{equation}
Thus, the corrected mixing operator retains unit row sums, but its entries
need not be nonnegative. No additional softmax is applied, preserving
the signed correction.
If the two operators commute, the correction vanishes and the message
reduces to $A_sV_n$. The zero row sums of the correction also imply that it
produces zero output when applied to a spatially constant value field.

After restoring the window messages to the spatial layout, we apply
GroupNorm, project them back to the stage channel width, and add a
channel-scaled residual:
\begin{equation}
  \Delta_s=\gamma_s\odot
  W_{\mathrm{out},s}\bigl(\GroupNorm(M_s)\bigr),\qquad
  x_s'=x_s+\Delta_s.
  \label{eq:stage-residual}
\end{equation}
The $1\times1$ projection $W_{\mathrm{out},s}$ maps 32 channels to the stage
channel width. Its weights are initialized from a truncated normal
distribution with standard deviation 0.02. The channelwise residual scaling
vector $\gamma_s$ is initialized to 0.1. The modified stage feature has the
same shape as its input and follows the routing in
Eq.~\eqref{eq:feature-routing}.

\subsection{Adaptive Multiscale Aggregation}

AMSA operates on the 1280-channel fused representation $z$. Three depthwise
$3\times3$ convolutions use dilation rates of 1, 2, and 3 to extract
responses over different spatial extents. Cross-scale weighting has also
been used in truncation compensation for high-resolution
segmentation~\cite{ref29}.
\begin{equation}
  u_d=\DWConv_{3\times3,d}(z),\qquad d\in\{1,2,3\}.
  \label{eq:dilated}
\end{equation}
A $1\times1$ convolution $G$ produces three scale scores at each pixel.
Softmax along the scale dimension yields $\alpha_d$; these weights are
shared across channels at each spatial position. Aggregation and residual
projection are defined as
\begin{equation}
  \alpha_d=\softmax_d(G(z)),\qquad
  u=\sum_{d=1}^{3}\alpha_d\odot u_d,
  \label{eq:scale-weights}
\end{equation}
\begin{equation}
  z'=z+\gamma_a\odot W_{\mathrm{proj}}(u).
  \label{eq:aggregation-residual}
\end{equation}
The dense $1\times1$ projection $W_{\mathrm{proj}}$ preserves the
1280-channel width. Its weights use truncated normal initialization with
standard deviation 0.02, and the channelwise scale $\gamma_a$ is initialized
to 0.1. The DANet prediction head receives $z'$.

\subsection{Supervision and Optimization}

\subsubsection{Loss and Optimization Strategy}

Training uses the baseline edge and mask Dice losses:
\begin{equation}
  \mathcal{L}=0.8\,\mathcal{L}_{\mathrm{Dice}}^{\mathrm{edge}}
  +0.16\,\mathcal{L}_{\mathrm{Dice}}^{\mathrm{mask}}.
  \label{eq:loss}
\end{equation}
The edge and manipulation-mask Dice losses flatten all pixels in
the local batch and reduce them jointly, using a smoothing constant of 1.
The Dice loss is
$1-(2\sum yp+1)/(\sum y^2+\sum p^2+1)$.
The single aggregated edge output and the mask output are each passed
through sigmoid before loss calculation and then weighted by 0.8 and 0.16,
respectively.

All new CBFI and AMSA parameters, including residual scales, mixing
coefficients, and log-temperatures, are assigned to a separate AdamW
parameter group with five times the learning rate of the shared parameters
and zero weight decay. Shared parameters use the same settings in both
models.

\subsubsection{Computational Complexity}

Let the spatial dimensions of stage $s$ be $h_s\times w_s$, with
$P=h_sw_s$, projected width $d$, and window length $L$. In CBFI, both
query--key multiplication and attention--value multiplication are constrained
to local windows and cost $\mathcal{O}(PLd)$. Composing the two attention
operators costs $\mathcal{O}(PL^2)$, and attention storage requires
$\mathcal{O}(PL)$. With a fixed window size, these costs grow linearly with
the spatial area of the stage. For AMSA, let $Q=hw$ and let $c$ be the
channel width. The three depthwise $3\times3$ convolutions have a total
cost of $\mathcal{O}(27Qc)$, while the dense output projection costs
$\mathcal{O}(Qc^2)$. For fixed $L$, $d$, and $c$, both modules scale
linearly with the number of spatial positions.

\begin{algorithm}[tbp]
  \caption{OCBI forward computation}
  \label{alg:forward}
  \begin{algorithmic}[1]
    \Require Image $I$; Co-Transformers components
    \State $x_0\gets I$
    \For{$s=1,\ldots,4$}
      \State $x_s\gets E_s(x_{s-1})$ \Comment{Complete encoding before fusion}
    \EndFor
    \State $n\gets\mathrm{CTE}(\mathrm{NoiseFusion}(I))$
    \For{$s=1,\ldots,4$}
      \State Resize $n$ to the spatial resolution of $x_s$ by bilinear interpolation
      \State Project both branches to 32 channels and partition into local windows
      \State Apply $\ell_2$ normalization to queries and keys
      \State Compute $A_s$, $N_s$, and $C_s=A_sN_s-N_sA_s$ in float32
      \State Compute $M_s=(A_s+\lambda_sC_s)V_{n,s}$ and restore the spatial layout
      \State $x_s'\gets x_s+\gamma_s\odot
        W_{\mathrm{out},s}(\GroupNorm(M_s))$
    \EndFor
    \State $\ell_e\gets H_{\mathrm{edge}}(x'_1,x'_2,x'_3,x'_4)$
    \State $z\gets\operatorname{Concat}(x'_4,n)$
    \State Compute dilated responses $u_d$ and pixelwise softmax weights $\alpha_d$
    \State $z'\gets z+\gamma_a\odot
      W_{\mathrm{proj}}(\sum_d\alpha_d\odot u_d)$
    \State $\ell_m\gets H_{\mathrm{mask}}(z')$
    \State $p\gets\sigma(\operatorname{Resize}(\ell_m,H,W))$;
      $e\gets\sigma(\operatorname{Resize}(\ell_e,H,W))$
    \State \Return Mask probability map $p$ and auxiliary edge map $e$
  \end{algorithmic}
\end{algorithm}

\FloatBarrier
\section{Experiments}

\subsection{Datasets and Splits}

We train on 4\,865 entries from CASIA v2.0 and use 258 entries for
validation. Table~\ref{tab:training-data} lists the training and test splits.
Counts refer to split-list entries, which may include duplicate images.

\begin{table}[tbp]
  \centering
  \caption{Training, validation, and test splits}
  \label{tab:training-data}
  \begin{tabular}{lrr}
    \toprule
    Split & List entries & Role \\
    \midrule
    CASIA v2.0 train & 4\,865 & task training \\
    CASIA v2.0 validation & 258 & monitoring \\
    CASIA v1.0 & 920 & clean test \\
    COVERAGE & 100 & clean test \\
    Columbia & 180 & clean test \\
    NIST16 subset & 240 & clean test \\
    IMD2020 & 2\,010 & clean test \\
    \bottomrule
  \end{tabular}
\end{table}

CASIA2 is split by source group with seed 0 into the training and validation
lists. Clean testing uses CASIA v1.0~\cite{ref7}, COVERAGE~\cite{ref19},
Columbia~\cite{ref10}, a fixed 240-entry NIST16 evaluation list, and
IMD2020~\cite{ref17}.
Evaluation lists preserve duplicate-content entries.

No identical paths or image hashes were found between the training and test
lists.
Masks are generally binarized at grayscale values
greater than 127, while Columbia uses a threshold greater than 99. NIST16
masks are inverted to mark manipulated regions. COVERAGE masks with
mismatched dimensions are aligned by nearest-neighbor interpolation.

\subsection{Training Settings}
\label{sec:training}

All experiments are conducted on four NVIDIA A800-SXM4-80GB GPUs with a
global batch size of 112. The image encoder is initialized from the
ImageNet-pretrained MiT-B3 model described for SegFormer~\cite{ref22}, while
the forensic branch uses the fixed noise-sensitive initialization of the
baseline extractor. We use AdamW with initial learning rates of
$1.37\times10^{-4}$ for shared parameters and $6.85\times10^{-4}$ for newly
introduced OCBI parameters. The learning rate is warmed up for two epochs and
then follows a cosine decay schedule. Both models are trained for 50 epochs.

Both models are trained on the CASIA-only split at $512\times512$ resolution
with the baseline mask construction and augmentation pipeline. The validation
split is used only for monitoring, and all reported clean and post-processing
results use the final epoch-50 model. The remaining optimization and execution
details are listed in Table~\ref{tab:training-config}.

\begin{table}[!ht]
  \centering
  \small
  \caption{Training configuration for retrained Co-Transformers and OCBI}
  \label{tab:training-config}
  \setlength{\tabcolsep}{5pt}
  \renewcommand{\arraystretch}{1.12}
  \begin{tabularx}{\linewidth}{@{}p{0.31\linewidth}Y@{}}
    \toprule
    \textbf{Setting} & \textbf{Configuration} \\
    \midrule
    Training data and input &
      CASIA-only training split; $512\times512$ inputs; baseline mask
      construction and geometric/photometric augmentation \\
    Objective &
      $0.8\mathcal{L}_{\mathrm{Dice}}^{\mathrm{edge}}+
       0.16\mathcal{L}_{\mathrm{Dice}}^{\mathrm{mask}}$ \\
    Optimizer &
      AdamW with $\beta_1=0.9$, $\beta_2=0.999$, and
      $\epsilon=10^{-12}$ \\
    Learning rate and regularization &
      Shared parameters: $1.37\times10^{-4}$, weight decay $0.05$;
      new OCBI parameters: $6.85\times10^{-4}$, weight decay $0$ \\
    Schedule &
      2-epoch warm-up followed by cosine decay to
      $5\times10^{-7}$ \\
    Hardware and batch size &
      Four NVIDIA A800-SXM4-80GB GPUs; per-device/global batch size
      $28/112$; no gradient accumulation \\
    Numerical precision &
      Mixed-precision training; CBFI attention computed in float32 \\
    Initialization &
      ImageNet-pretrained MiT-B3 image encoder; fixed noise-sensitive
      forensic branch \\
    Training budget &
      50 epochs, 2\,200 optimizer updates, random seed 42 \\
    Evaluation checkpoint &
      Validation for monitoring only; final epoch-50 model for clean and
      post-processing evaluation \\
    \bottomrule
  \end{tabularx}
\end{table}

\subsection{Evaluation Protocol}

Evaluation uses $512\times512$ inputs and float32 inference. Manipulation
probabilities are binarized at a threshold of 0.5. Pixelwise F1 for
a single image is defined as
\begin{equation}
  F_1=\frac{2\mathrm{TP}}{2\mathrm{TP}+\mathrm{FP}+\mathrm{FN}},
  \label{eq:f1}
\end{equation}
where TP, FP, and FN denote the numbers of true-positive, false-positive,
and false-negative pixels, respectively. Dataset-level PixelF1 is the
arithmetic mean of per-image F1 scores. The primary metric is the equally
weighted mean of the five test-dataset scores:
\begin{equation}
  F_{\mathrm{macro}}=\frac{1}{5}\sum_{d=1}^{5}F_{1,d}.
  \label{eq:macro}
\end{equation}
Each test image is evaluated exactly once, without distributed-sampling
padding. Manipulation masks are nonempty for all images in the test lists
used.

\subsection{Main Comparison}

Table~\ref{tab:main} reports localization on clean test images.
OCBI achieves higher PixelF1 on
all five test datasets, while the baseline has higher macro AUC.
PixelF1 measures mask accuracy at the fixed threshold; AUC measures
ranking performance across thresholds.

\begin{table}[!htbp]
  \centering
  \small
  \caption{Clean localization at the final checkpoint}
  \label{tab:main}
  \begin{tabular}{@{}lrrrr@{}}
    \toprule
    Dataset & Co-Transformers F1 & OCBI F1 & Co-Transformers AUC & OCBI AUC \\
    \midrule
    CASIA v1.0 & 0.7409 & \textbf{0.7452} & 0.9536 & \textbf{0.9572} \\
    COVERAGE & 0.2853 & \textbf{0.2994} & \textbf{0.8868} & 0.8826 \\
    Columbia & 0.8591 & \textbf{0.8613} & 0.9288 & \textbf{0.9313} \\
    NIST16 subset & 0.5605 & \textbf{0.5607} & \textbf{0.9072} & 0.8928 \\
    IMD2020 & 0.4192 & \textbf{0.4197} & 0.8641 & \textbf{0.8724} \\
    Macro average & 0.5730 & \textbf{0.5773} & \textbf{0.9081} & 0.9073 \\
    \bottomrule
  \end{tabular}
  \par\smallskip
  \begin{minipage}{\linewidth}\footnotesize
    Higher is better. Bold indicates the higher score within each
    metric pair. F1 is computed at threshold 0.5; AUC is averaged over valid
    image-level rankings. Macro averages weight the five datasets equally.
  \end{minipage}
\end{table}

\begin{figure}[!htbp]
  \centering
  \includegraphics[width=\linewidth]{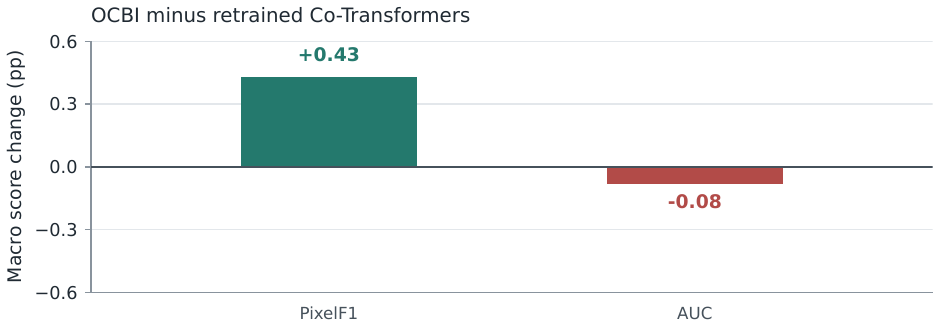}
  \caption{Changes in five-dataset macro PixelF1 and AUC relative to the
  retrained Co-Transformers baseline. Bars show signed changes in the
  equally weighted macro scores, expressed in percentage points (pp), defined as
  $100\times(\mathrm{OCBI}-\mathrm{baseline})$.
  The difference axis is symmetric about zero.}
  \label{fig:endpoints}
\end{figure}

Figure~\ref{fig:endpoints} shows the aggregate changes relative to the
retrained baseline: macro PixelF1 increases
by 0.43 percentage points, while macro AUC decreases by 0.08 percentage
points. On the validation split, OCBI reaches F1 0.6240 and AUC 0.9391,
compared with 0.6074 and 0.9298 for the baseline.

\begin{figure}[!htbp]
  \centering
  \includegraphics[width=\linewidth]{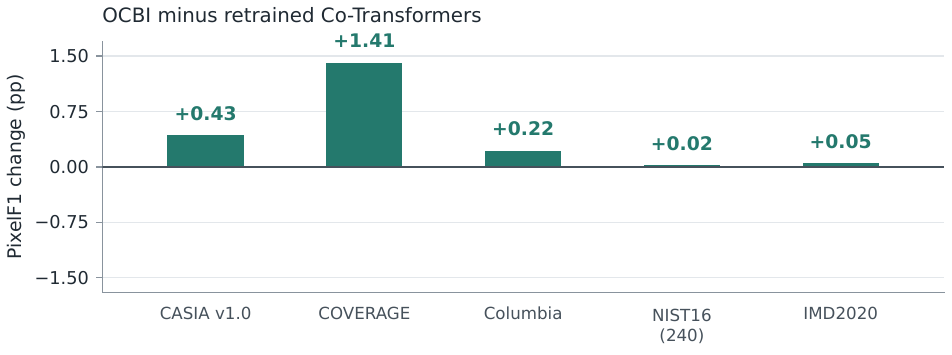}
  \caption{Dataset-level changes in clean localization performance.
  Bars show OCBI minus retrained Co-Transformers PixelF1 in percentage
  points (pp) on a symmetric zero-centered axis. Labels quantify the
  change for each dataset; absolute scores are reported in
  Table~\ref{tab:main}.}
  \label{fig:dataset-differences}
\end{figure}

\begin{samepage}
Figure~\ref{fig:dataset-differences} shows the signed PixelF1 differences
across the five test datasets. Gains range from 0.02 percentage points
on NIST16 to 1.41 percentage points on COVERAGE. AUC increases on
CASIA v1.0, Columbia, and IMD2020, and decreases on COVERAGE and NIST16.

\end{samepage}

\FloatBarrier

\subsection{Robustness to Post-Processing}

We evaluate JPEG compression at quality levels 90, 70, and 50; Gaussian
blur with standard deviations 1 and 2; and resizing factors of 0.75 and
0.5 on CASIA v1.0, COVERAGE, and Columbia. The same transformation
pipeline is used for both models, with input resizing enabled.
Scaling uses bilinear interpolation.

\begin{table}[!htbp]
  \centering
  \small
  \caption{PixelF1 under post-processing}
  \label{tab:robustness}
  \setlength{\tabcolsep}{4pt}
  \renewcommand{\arraystretch}{1.2}
  \begin{tabularx}{\linewidth}{@{}p{0.23\linewidth}*{6}{>{\centering\arraybackslash}X}@{}}
    \toprule
    \textbf{Transformation} & \multicolumn{2}{c}{\textbf{CASIA v1.0}} &
      \multicolumn{2}{c}{\textbf{COVERAGE}} &
      \multicolumn{2}{c}{\textbf{Columbia}} \\
    \cmidrule(lr){2-3}\cmidrule(lr){4-5}\cmidrule(lr){6-7}
    & Baseline & OCBI & Baseline & OCBI & Baseline & OCBI \\
    \midrule
    JPEG Q90 & 0.7035 & \textbf{0.7073} & 0.1965 & \textbf{0.1996} & 0.8179 & \textbf{0.8402} \\
    JPEG Q70 & \textbf{0.6126} & 0.5992 & \textbf{0.2055} & 0.1707 & 0.6692 & \textbf{0.7448} \\
    JPEG Q50 & \textbf{0.4885} & 0.4746 & \textbf{0.1906} & 0.1422 & 0.5767 & \textbf{0.6709} \\
    \addlinespace[3pt]
    Blur $\sigma=1$ & 0.6577 & \textbf{0.6694} & \textbf{0.1601} & 0.1264 & 0.7474 & \textbf{0.7681} \\
    Blur $\sigma=2$ & 0.5749 & \textbf{0.5767} & 0.1432 & \textbf{0.1768} & 0.5810 & \textbf{0.6089} \\
    \addlinespace[3pt]
    Scale 0.75 & 0.6731 & \textbf{0.6788} & \textbf{0.1626} & 0.1600 & 0.7979 & \textbf{0.8086} \\
    Scale 0.5 & 0.6271 & \textbf{0.6400} & \textbf{0.1338} & 0.1177 & 0.6821 & \textbf{0.7203} \\
    \bottomrule
  \end{tabularx}
  \par\smallskip
  \begin{minipage}{\linewidth}\footnotesize
    CASIA denotes CASIA v1.0. Baseline denotes retrained Co-Transformers.
    Bold indicates the higher PixelF1 within each dataset and transformation.
    Blur values are Gaussian standard deviations; scale values are scaling
    factors.
  \end{minipage}
\end{table}

Across the 21 dataset--transformation conditions, OCBI improves PixelF1
in 14 conditions, raising the equally weighted average from 0.4953 to
0.5048. On Columbia, it improves under all seven transformations, with
the largest gain of 9.42 percentage points at JPEG Q50. On CASIA v1.0,
it improves under both blur levels and both resizing factors.

\FloatBarrier
\subsection{Analysis and Discussion}

In the CASIA-only comparison with retrained
Co-Transformers~\cite{ref23}, OCBI improves PixelF1 on all five clean test
datasets, increasing the macro score from 0.5730 to 0.5773, a gain of
0.43 percentage points. The largest improvement is on COVERAGE, at
1.41 percentage points (Table~\ref{tab:main}). The gains across datasets
indicate that OCBI improves
thresholded localization beyond the training dataset.

We also tested JPEG compression, Gaussian blur, and resizing on three
datasets. Table~\ref{tab:robustness} shows improvements in 14 of the
21 conditions, with aggregate PixelF1 rising from 0.4953 to 0.5048,
or 0.95 percentage points. Columbia improves under every tested
transformation, including a gain of 9.42 percentage points at JPEG Q50.
These improvements extend
to compressed, blurred, and resized inputs, consistent with its use of
local forensic relationships and multiple spatial contexts.

\section{Conclusion}

We proposed OCBI, a Co-Transformers extension for image manipulation
localization. The method models the difference between two compositions
of local semantic and forensic mixing operators through a signed
commutator correction and combines spatial contexts through pixelwise
adaptive multiscale aggregation.

OCBI improves localization over the retrained baseline on
the clean test datasets and under a range of image post-processing
conditions. The experiments support using composition-order differences
and adaptive spatial aggregation to combine semantic and forensic
evidence for manipulated-region prediction.


\end{document}